\documentclass[letterpaper, 10 pt, conference]{ieeeconf}  % Comment this line out if you need a4paper

\usepackage{graphics}
\usepackage{upgreek}
\usepackage{amssymb}
\usepackage{amsfonts}
\usepackage{amsmath}
\usepackage{refstyle}
\usepackage{xcolor}
\usepackage{subfigure}
\usepackage{todonotes}
\usepackage{mathtools}
\usepackage{algorithm}
\usepackage{algpseudocode}
\usepackage{cite}
\usepackage{multirow}
\usepackage[para,online,flushleft]{threeparttable}

\usepackage{hyperref}

\newcommand\copyrighttext{%
  \footnotesize \textcopyright 2022 IEEE. Personal use of this material is permitted.
  Permission from IEEE must be obtained for all other uses, in any current or future
  media, including reprinting/republishing this material for advertising or promotional
  purposes, creating new collective works, for resale or redistribution to servers or
  lists, or reuse of any copyrighted component of this work in other works.
  DOI: \href{https://ieeexplore.ieee.org/document/10011794}{No. 10.1109/ROBIO55434.2022.10011794}}
\newcommand\copyrightnotice{%
\begin{tikzpicture}[remember picture,overlay]
\node[anchor=south,yshift=10pt] at (current page.south) {\fbox{\parbox{\dimexpr\textwidth-\fboxsep-\fboxrule\relax}{\copyrighttext}}};
\end{tikzpicture}%
}

\IEEEoverridecommandlockouts                              % This command is only needed if
\title{\LARGE \bf
Towards Precise Model-free Robotic Grasping\\ with Sim-to-Real Transfer Learning}
\author{Lei Zhang$^{1,2}$\dag, Kaixin Bai$^{1,2}$\dag, Zhaopeng Chen$^{2,1}$*, Yunlei Shi$^{1,2}$, Jianwei Zhang$^{1}$% <-this % stops a space
\thanks{\dag The first two authors contribute equally to this paper.}% <-this % stops a space
\thanks{*Corresponding author.}
\thanks{{$^{1}$TAMS (Technical Aspects of Multimodal Systems), Department of
Informatics, Universit\"at Hamburg}, {$^{2}$Agile Robots AG}}
}
\DeclarePairedDelimiterX{\norm}[1]{\lVert}{\rVert}{#1}
\begin{document}

\maketitle
\copyrightnotice

\thispagestyle{empty}
\pagestyle{empty}
% @@@@@@@@@@@@@@@@@@@@@@@@@@@@@@@@@@@@@@@@
\begin{abstract}    
Precise robotic grasping of several novel objects is a huge challenge in manufacturing, automation, and logistics. Most of the current methods for model-free grasping are disadvantaged by the sparse data in grasping datasets and by errors in sensor data and contact models. This study combines data generation and sim-to-real transfer learning in a grasping framework that
reduces the sim-to-real gap and enables precise and reliable model-free grasping. A large-scale robotic grasping dataset with dense grasp labels is generated using domain randomization methods and a novel data augmentation method for deep learning-based robotic grasping to solve data sparse problem. We present an end-to-end robotic grasping network with a grasp optimizer. The grasp policies are trained with sim-to-real transfer learning. % based 
The presented results suggest that our grasping framework reduces the uncertainties in grasping datasets, sensor data, and contact models. In physical robotic experiments, our grasping framework grasped single known objects and novel complex-shaped household objects with a success rate of 90.91\%. In a complex scenario with multi-objects robotic grasping, the success rate was 85.71\%. The proposed grasping framework outperformed two state-of-the-art methods in both known and unknown object robotic grasping.
\end{abstract}

\section{Introduction}
\label{intro}
From a perception viewpoint, grasping is among the most important and challenging control tasks in robotic
manipulation. A grasping task requires critical empirical knowledge and skills from human beings. 
However, sensor noise and control imprecision continue to limit the robustness of robotic grasping in cluttered scenes~\cite{mahler2018dex, mahler2019learning, staub2019dex}.

Unlike traditional model-based robotic grasping based on six-dimensional (6D) pose estimation, model-free grasping approaches do not rely on the availability of the a three-dimensional (3D) object model~\cite{Carbone:2012:GR:2422908,du2019vision}. To grasp unknown objects, developers have built robust methods for model-free grasping, including analytical approaches, data-driven methods, deep learning-based methods, and others~\cite{Review_roboticgraspdetection2018,Carbone:2012:GR:2422908,du2019vision,mahler2017dex,mahler2019learning,staub2019dex,zeng2018robotic}.% 

Analytical methods have shown that a grasp is unstable without force closure when it violates the antipodal condition~\cite{nguyen1987constructing,ding2000efficient,sahbani2012overview}. Physical concepts such as the object wrench space, grasp wrench space (GWS)~\cite{borst2004grasp}, and Ferrari–Canny metric~\cite{canny1992planning} have been proposed for estimating the stable grasp candidates and grasp quality~\cite{sahbani2012overview,miller2003automatic}. The GWS calculation uses discretization methods in an accelerated algorithm, which may increase the error of the contact-model calculation. Meanwhile, the robot grasping based on 6D pose estimation is affected by errors in the sensor noise and discretization methods.

Leveraging data-driven methods, convolutional neural networks (CNNs) learn the grasping of novel objects on datasets from knowledge of human nature~\cite{zeng2020tossingbot,bohg2013data,zeng2018robotic,mahler2017dex,staub2019dex}. However, high-precision empirical methods aimed at performing precise model-free grasping tasks are challenged by the limited number of objects and grasping labels in grasping databases. When the geometric features of the objects to be grasped in the home differ from those of the objects in the robot grasping datasets, the performance of the neural network for model-free robotic grasping is degraded. Simulation grasping datasets using known grasp-sampling methods and grasp quality metrics have been recently proposed for neural network training. Examples are the Jacquard grasping dataset~\cite{depierre2018jacquard}, the grasp dataset in Dex-Net 2.0~\cite{mahler2017dex} and GraspNet-1Billion~\cite{fang2020graspnet}. However, in the current large-scale robotic grasping datasets, the grasp candidates are represented only as single pixels describing their grasp location and orientation in the image; consequently, the pixel grasping labels are sparsely distributed in the dataset. Furthermore, the labels in the generated image datasets are imbalanced. These problems affect the convergence of the neural network weights. In our qualitative study of robotic grasping methods with an image-wise neural network, we found that when the final grasp candidates are selected from the grasp affordance map, they sometimes violate the force closure condition or yield inferior grasp-quality metrics (see Appendix~\ref{sec:appendix_qualitative_study}). For this reason, database generation by human annotation has earned a bad reputation for low speed and unstable results~\cite{du2019vision,zeng2018robotic}. % 

Herein, we propose a robust and productive model-free robotic grasping framework, which overcomes the above limitations by combining a data generation approach and sim-to-real transfer learning methods. Our framework
increases the reliability of model-free grasping. 
Our core contributions are summarized below.
\begin{enumerate}
\item Using a novel data augmentation, our data generation process creates a large-scale robotic grasping dataset with dense grasp labels.
\item The grasp affordance maps are learned from depth images using sim-to-real transfer learning methods achieving accurate model-free robotic grasping.
\item By considering the force-closure metric and gripper model in the grasp optimizer, we reduce the uncertainties and enhance the performance and robustness of the framework.
\item We combine \textbf{1)}, \textbf{2)} and \textbf{3)} into a grasping framework and compare its performance with those of two state-of-the-art methods~\cite{zeng2018robotic,2019Metagrasp} in a qualitative study and real-world experimental setups.
\item The success rate was 90.91\% for grasping of single known objects and novel household objects with complicated shapes, and 85.71\% for multi-objects robotic grasping in a complex scenario.
\end{enumerate}

The remainder of this paper is divided into five sections. 
Section~\ref{sec:relatedwork} summarizes related work and Section~\ref{sec:method} describes the proposed data generation and robotic grasping frameworks. Section~\ref{sec:experiment} provides the experimental framework and qualitatively evaluates the proposed methods. The results are presented in Section~\ref{sec:results}. Section~\ref{sec:conculsion} summarizes the paper and mentions future possibilities. 
\section{Related Work}
\label{sec:relatedwork}
\subsection{Analytical Methods for Grasping}
A stable grasp requires force closure to constrict the movement of the grasped object~\cite{murray2017mathematical}\cite{nguyen1987constructing}. The GWS, which is related to the force and torque of the friction cone, refers to the affordability of the grasp points of the grasped object~\cite{borst2004grasp}.

To evaluate the grasp quality, researchers have calculated the GWS using different measurement methods~\cite{borst2004grasp}. For instance, the German Aerospace Center estimated the GWS of friction cones using a discretization method, and hence compiled a stable grasping planner~\cite{borst1999fast}. However, discretization largely contributes to error in the GWS approximation. To avoid this error, Ferrari and Canny calculated the grasp quality based on the ability of contact wrenches to withstand 
maximum disturbances~\cite{canny1992planning,borst2004grasp}. The Ferrari–Canny metric is now commonly employed in selections of stable grasp candidates~\cite{lin2016taskgrasp,mahler2017dex}.

\subsection{Deep Learning-based Methods For Robotic Grasping}
Deep learning-based methods such as supervised learning methods and reinforcement methods are rapidly replacing 6D
pose estimation for robotic grasping~\cite{du2019vision}.

Mahler et al.~\cite{mahler2017dex} presented the robust robotic grasping framework Dex-Net 2.0 with a Grasp Quality Convolutional Neural Network, which estimates the success probability of grasping with parallel-jaw grippers. For this purpose, they generated a database of more than 500,000 images containing the local contact features of grasping. Dex-Net 2.0 efficiently performs robotic grasping tasks despite lacking the global features of the objects and requiring the adjustment of many parameters. These disadvantages reduce the flexibility and universality of Dex-Net 2.0.

Pixel-wise deep learning methods predict the grasp quality, gripper width, and grasp direction in each pixel of an input image. These methods also achieve high success rates of robotic grasping~\cite{zeng2018robotic,morrison2018closing,asif2019densely}, but their performances are degraded by sparse grasping pixel labels and errors in current grasping datasets.

\subsection{Grasping Datasets and Data Generation Methods}
For the 2017 Amazon Robotics Challenge, Zeng et al.~\cite{zeng2018robotic} built a grasping dataset of household objects with human annotation and trained a pixel-wise network for suction and grasping. Nonetheless, establishing a large dataset with reliable human annotations is both time-consuming and expensive.

To overcome the limitations of the human-annotated datasets, many researchers
apply data generation methods that generate a reliable robotic grasping dataset~\cite{yan2019data,mahler2017dex,2019Metagrasp}. The Jacquard grasping dataset was constructed in simulation environments~\cite{depierre2018jacquard} and the grasp image dataset of Dex-Net 2.0~\cite{mahler2017dex} was constructed using analytical methods, including random grasp sampling and grasp quality metrics. The data samples in Dex-Net 2.0 are sensor data with local features near each grasp candidate. The dataset of MetaGrasp was built by trial-and-error in the V-REP Simulator~\cite{2019Metagrasp}. However, as the current datasets label only the grasp central points on the object, they are data-sparse. This problem increases the uncertainty of the current pixel-wise deep learning-based grasping methods.
%However

\subsection{Sim-to-Real Transfer Learning in Robotic Manipulation}
Sim-to-real transfer is thought to reduce the gap between the datasets and real situations and improve the performance of dexterous robotic manipulation methods~\cite{ho2020retinagan,zhao2020sim,pashevich2019learning}. In sim-to-real transfer, domain randomization can effectively randomize the synthetic parameters and simulate the real-world data distribution~\cite{alghonaim2020benchmarking,bousmalis2018using,james2017transferring,tobin2017domain}.

\section{Problem Statement and Method}\label{sec:method}
The core methods in our robotic-grasping framework are summarized in Fig.~\ref{fig:pipeline_overview} and outlined in the following sections. The framework consists of the following parts:
\begin{itemize}
\item Characterize the model database of robotic grasping and generate a large-scale grasping database using \textbf{domain randomization methods}.
\item Present a novel \textbf{data augmentation} for generating the synthetic grasping datasets.
\item \textbf{Sim-to-real transfer learning} of the encoder-decoder network for robotic grasping using Intel Realsense camera with complicated noise.
\item Build a grasping execution system with the trained network and the proposed \textbf{grasp optimizer}.
\end{itemize}
\subsection{Problem Statement}
Our main intention was to build a robotic grasping framework with a parallel-jaw gripper, which operates robustly and precisely despite uncertainties in the sensor data and computational contact models. We also attempt to solve the sparse grasping data problem of current grasping datasets. Assuming uniform mass distribution of the grasped object, we defined our problem as follows: %

\subsubsection{\textbf{Grasp Representation}}

A grasp pose is defined as a 3D position oriented along the Z-axis of the end-effector. Let $\boldsymbol{v}=(x_{\mathrm{grasp}}, y_{\mathrm{grasp}},z_{\mathrm{grasp}},$ $\theta_{\mathrm{grasp}})$ refer to the four degrees-of-freedom (4DOF) grasp pose, especially, \(\theta_{\mathrm{grasp}}\).

\subsubsection{\textbf{Grasp Affordance Map}}

Employing the neural network, we predict a grasp affordance map \(\boldsymbol{I}^{\mathrm{Affordance Map}}\). Each pixel in the grasp affordance map is the grasp central point of the quality distribution of the grasp. 

\begin{figure*}[htbp]
\centering
    \includegraphics[height=5.5cm]{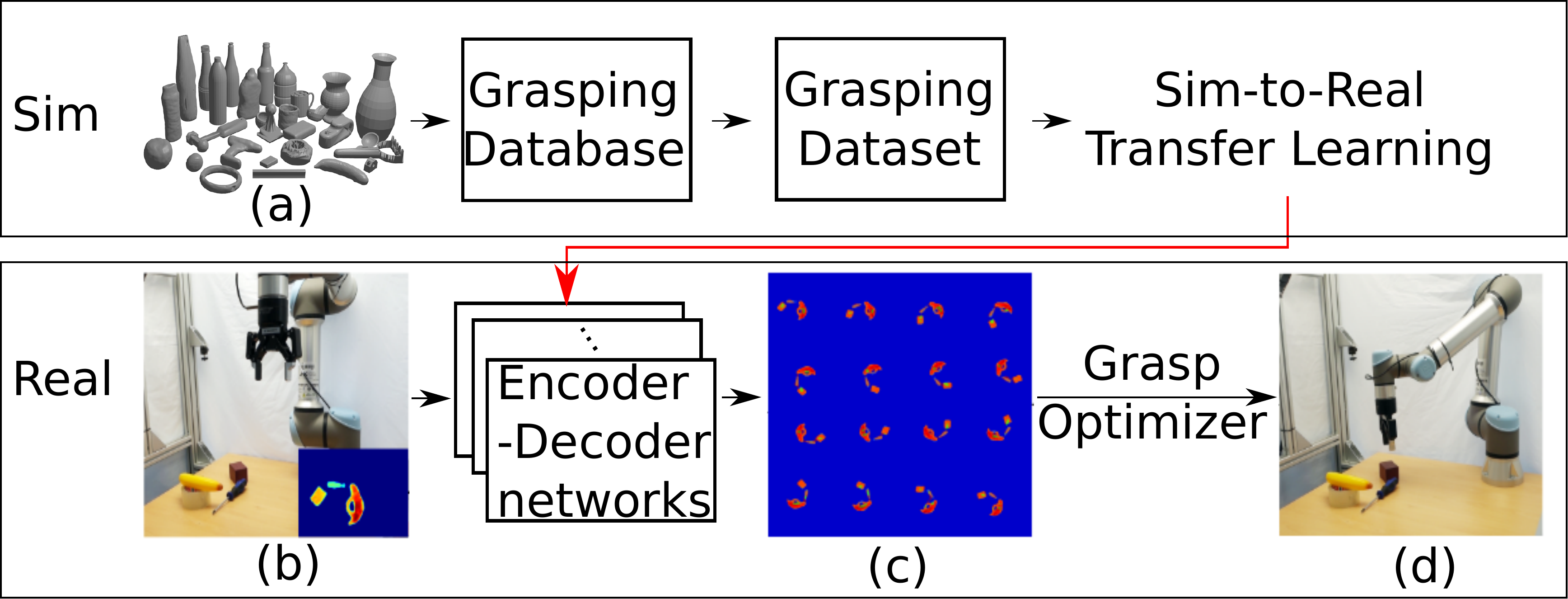}
\centering
\caption{Our proposed robotic grasping framework: (a) Based on the model database, generate a grasp database with horizontal grasp samples and grasp qualities. Next, build a grasping dataset of depth images and grasp affordance maps using our data augmentation and grasp orientation quantization; (b) input the depth images to the encoder-decoder grasp affordance network; (c) Grasp affordance maps. (d) using the grasp optimizer, select and execute an optimal grasp candidate.}
\label{fig:pipeline_overview}
\end{figure*}

\begin{figure}[htbp]
\centering
\includegraphics[width=8.5cm]{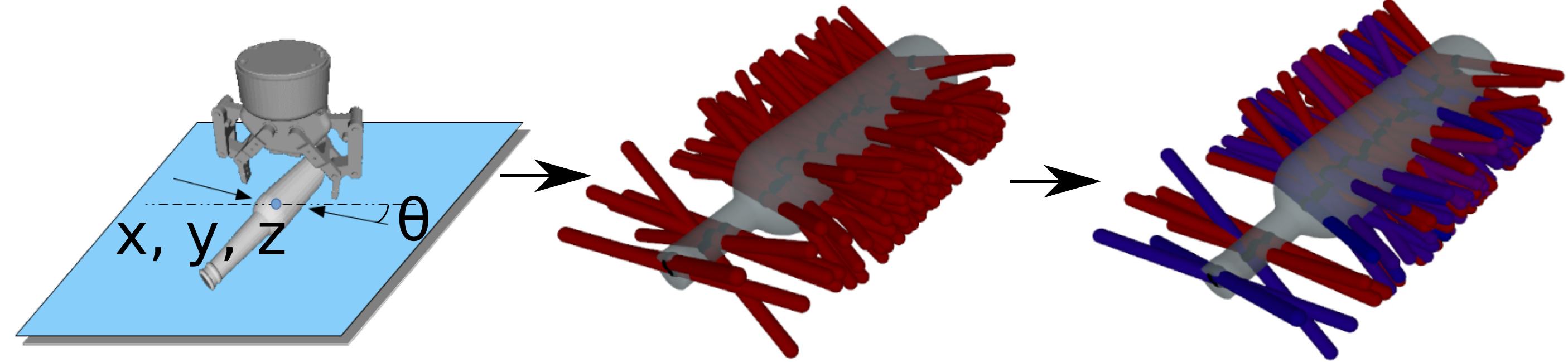}
\caption{Generation pipeline of horizontal grasp samples: (left) Example of a 4DOF grasp pose; (middle) only grasps oriented approximately perpendicular to the table are considered for grasp generation; (right) each grasp is rated between 0 and 1 inclusive based on the Ferrari-Canny metric (red: high-quality grasp, blue: low-quality grasp).}
\label{fig:data_generation_perpendicular_grasp}
\end{figure}

\subsubsection{\textbf{Encoder-decoder Grasp Affordance Network}}~

We develop an encoder-decoder grasp affordance network that learns a robust robotic grasp policy \(I^{\mathrm{grasp\ affordance\ map}} = \pi^{\mathrm{encoder\ decoder\ network}}(I^{\mathrm{depth\ image}})\). %

\subsection{Building a Model Database}
To train a flexible and robust CNN for robotic grasping, we generated a large dataset in a simulation environment. The model database with 1,521 commonly used object models was collected from 3dNet \cite{wohlkinger20123dnet}, the Yale-CMU-Berkeley (YCB) Dataset \cite{calli2017yale}, Princeton ModelNet \cite{wu20153d}, data models from Dex-Net \cite{mahler2017dex} \cite{mahler2019learning}, and the MVTec Industrial 3D Object Detection Dataset (MVTec ITODD) \cite{drost2017introducing}. All models were divided into four shape categories: spheroidal, cuboidal cup-like, and complicated. %

\subsection{Generating a Grasping Database with Domain Randomization}
To build a model-free grasping network, we must generate a grasping dataset with adequate data diversity and a distribution that covers the real-world data.

To address this problem, we first generate the grasp scenes of multifarious 3D models in a pyrender simulation environment~\cite{pyrender} with \textbf{domain randomization}. In the simulation environment, the grasped objects are randomly placed on a planar table and a synthetic camera is placed from the table at a random separation in the range [65 mm, 75 mm]. Object volumes are randomly generated in the range \([27~\text{cm}^{3}, 1000~\text{cm}^{3}]\), the typical volume distribution of daily household items. During network training, random noise with a normal distribution (mean = 1, standard deviation = 0.01) is added to the depth image input. Within the simulation environmental setup, the depth images of grasp scenes are rendered and added to the grasping database.
\begin{figure}[htbp]
\centering
\includegraphics[width=8.5cm]{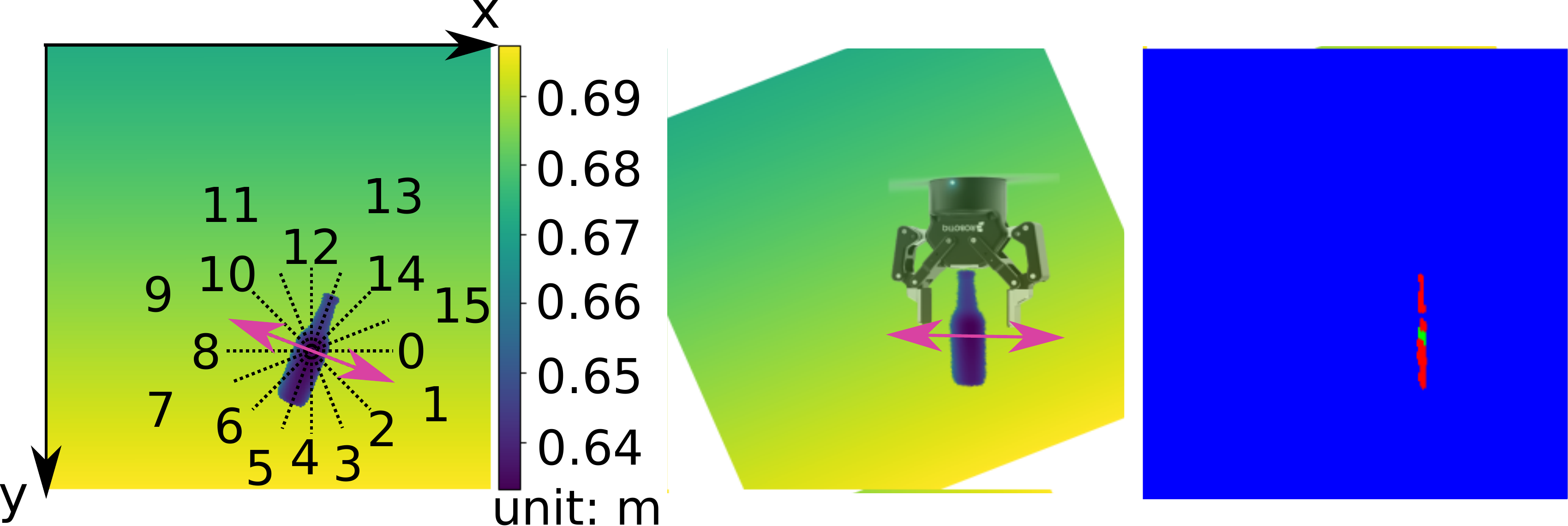}
\caption{Example of grasp orientation quantization: (left) The grasp orientation is quantized by the angle between the grasp orientation and the x-axis of the image coordinate system; (middle) oriented rendered depth map of the first or ninth quantized orientations; (right) grasp affordance map of grasp samples in the first or ninth quantized orientations.}
\label{fig:orientation_quantization_example}
\end{figure}

After obtaining the grasp scenes, horizontal grasp samples are generated under the force-closure grasp policy. This step must consider the poses of the object on the table. Similarly to related works on robotic grasping~\cite{zeng2018robotic,mahler2017dex}, only horizontal grasps generated approximately perpendicular to the planar surface are considered as grasping solutions, as shown in Fig.~\ref{fig:data_generation_perpendicular_grasp}(middle panel). In each scenario, 150 horizontal grasp samples are sampled and added to the grasp database. Grasping scenes with insufficiently many grasp samples are ignored. The proposed horizontal grasp sample is represented by the position of the grasp central point, the orientation angle $\theta$ between the grasp direction, and the x-axis of the image coordinate system. For evaluating the grasp samples, we calculate the GWS of the grasp candidates and estimate the qualities of the grasp candidates using the  Ferrari-Canny metric~(Fig.~\ref{fig:data_generation_perpendicular_grasp}, right panel). The final grasp database includes the rendered depth images, horizontal grasp candidates, and their corresponding estimated grasp qualities.

Finally, the labels of the grasping database are generated. To this end, our framework rotates the original image input toward the grasp orientation, thus avoiding the need to predict the grasp orientation. The horizontal grasps are then estimated. The orientations of grasp samples are quantized into 16 groups based on the angle $\theta$ between the grasp orientation and the x-axis of the image coordinate system (see Fig.~\ref{fig:orientation_quantization_example}). After \textbf{grasp orientation quantization}, the grasp orientations are specified by numbers in the range 0-15 and the rendered depth images are rotated until
their grasp orientations are parallel to the x-axis of the image coordinate system(see Fig.~\ref{fig:orientation_quantization_example}, middle panel). 

\begin{figure}[htbp]
\centering
\includegraphics[width=8cm]{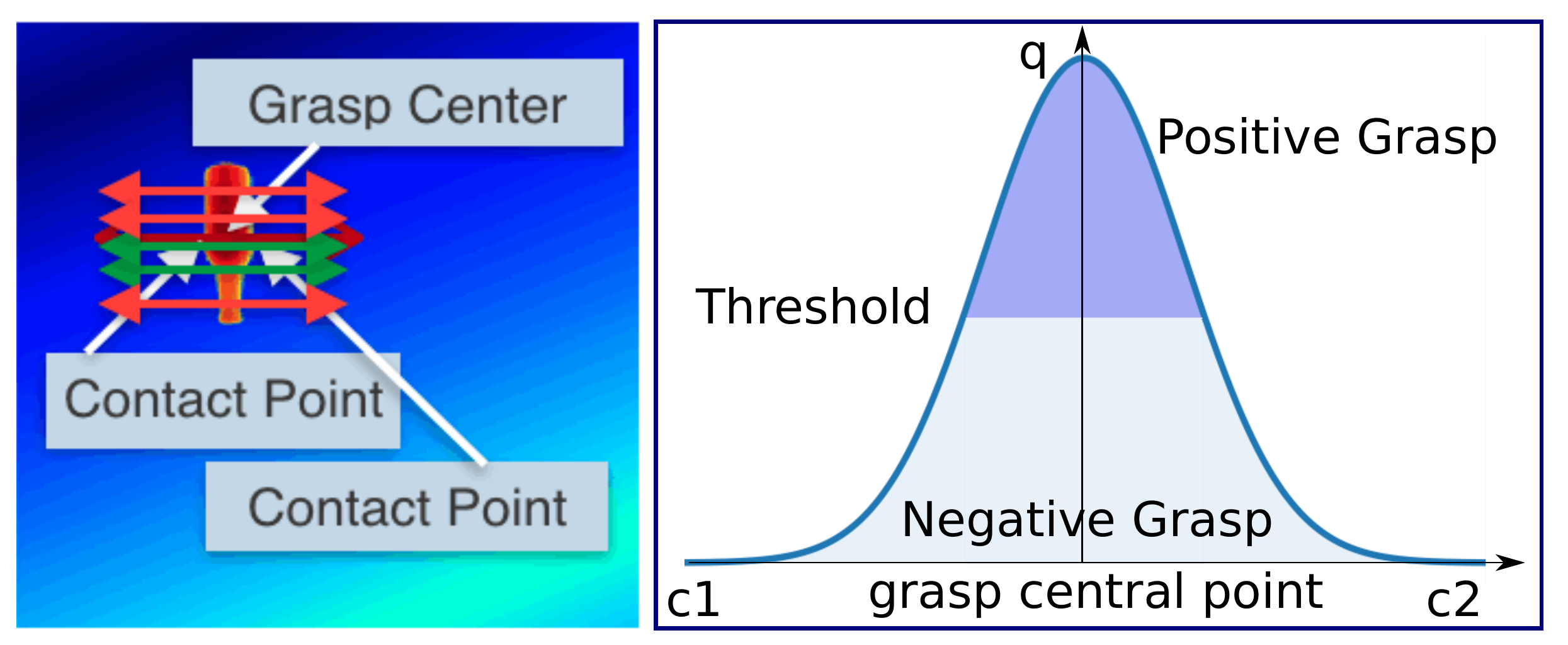}
\caption{Data augmentation between the grasp contact points. (c1: left grasp contact point; c2: right grasp contact point; g: grasp center; q: grasp quality.)}
\label{fig:gaussian_augmentation_data_augmentation_def}
\end{figure}
\begin{figure}[htbp]
\centering
\includegraphics[width=7.5cm]{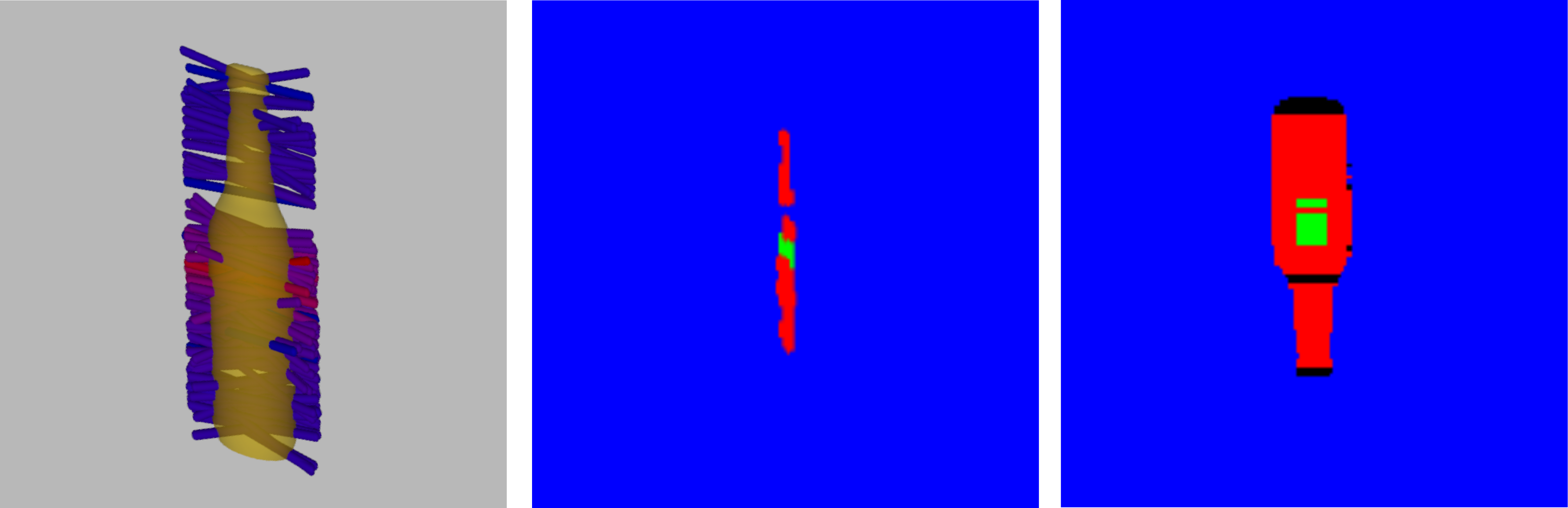}
\caption{Grasp affordance maps without and with data augmentation; (left) Grasp candidates in the scenario; (middle and right) affordance maps without and with the data augmentation.}
\label{fig:gaussian_augmentation_data_augmentation_comparison}
\end{figure}

The next task generates the grasp affordance label maps based on the horizontal grasp samples and their corresponding quantized orientations (Fig.~\ref{fig:orientation_quantization_example}, right panel). The grasp affordance maps are generated as labels of the encoder-decoder grasping network. Based on the normalized grasp qualities in the grasp data database, the grasp candidates are then labeled into three classes: red (negative grasps with higher grasp qualities), green (positive grasps with lower grasp qualities), and blue (background). We also defined a quality threshold \(\theta_q\) as the median value of the grasp qualities. The positive-grasp samples represent the candidates with qualities exceeding \(\theta_q\); the remainder are classified into the negative-grasp class. In the grasp
affordance map, the central points of all grasp samples are labeled within their grasp classifications (red for negative grasps, green for positive grasps, and blue for background pixels).

\subsection{Augmenting Grasping Labels with Distribution Functions and Grasp Qualities.}
The recent open-source grasping datasets~\cite{depierre2018jacquard,mahler2017dex,2019Metagrasp} suffer from the sparsity problem of their grasp pixel labels. After training the grasp networks on a synthetic database with the same format as the open-source grasping datasets, we found that the sparsity problem causes the estimation of false grasp candidates at the edge of the grasped object or on the table (see Fig.~\ref{fig:inference_map}). The grasping network performs poorly because the grasps are represented by an insufficient number of pixel labels. To
resolve this problem, we propose a \textbf{data augmentation method} that boosts the quantity and diversity of horizontal grasp samples and increases the number of grasp labels according to the following function:

\begin{equation}
q(x)=e^{-\left(\frac{x-g}{2}\right)^{2}} q\_original(g) 
\end{equation}
where, $x$ represents the pixel position sampled between contact points, $q\_original(g)$ represents the quality of the corresponding grasp center, and $q(x)$ denotes the augmented quality, as shown in Fig.~\ref{fig:gaussian_augmentation_data_augmentation_def}. Using our data augmentation method, we classify the augmented grasp candidates and generate an augmented grasp affordance map (Fig.~\ref{fig:gaussian_augmentation_data_augmentation_comparison}, right panel). Based on our qualitative study in Appendix~\ref{sec:appendix_statistical_analysis}, our augmentation method increases the number of grasp labels between the left and right contact points and reduces the uncertainty of the network, thus avoiding the prediction of positive grasps on the object edge.

Finally, a synthetic grasping dataset is generated from the rendered depth images and grasp affordance label maps with data augmentation.

\subsection{Learning the encoder-decoder grasp affordance network with Sim-to-Real Transfer}
For model-free grasping, we developed a deep residual encoder-decoder network that predicts the grasp affordance maps and the possibilities of horizontal robotic grasps from depth images (which are input data to the network). The encoder in our network includes the layers of the ResNet-50 v2 network \cite{he2016identity} without the fully connected layers at the end. The ResNet network is trained by transfer learning based on the ImageNet pre-trained model~\cite{deng2009imagenet}. The decoder part consists of convolutional, batch normalization, and bilinear upsampling layers. The input layer, which is sized to fit the image size ($300 \times 300$ pixels) with three channels, accepts the depth image as input. The  final layer is a ($300 \times 300 \times 3$)-sized grasp affordance map encoding the possibility of a robotic horizontal grasp at each pixel. The three channels of the output data represent the possibilities of the three classes (negative and positive grasp central points in red and , respectively, and the blue
background). 

In grasp affordance label maps, the very different numbers of the three classes are problematic for network training. Such label imbalance leads to the dilemma of training with a normal cross-entropy loss function. Here we formulate the following \textbf{adaptable weight cross-entropy loss function}:
\begin{equation}
\centering
\begin{aligned}
    L=-\frac{1}{3 H W} \sum_{i=1}^{H} \sum_{j=1}^{W} \sum_{k=1}^{3} \boldsymbol{M}_{i j k}^{\mathrm{adaptive}} \boldsymbol{\Psi}_{i j k}  \log \frac{e^{\Theta(\boldsymbol{x})_{i j k}}}{\sum_{l=1}^{3} e^{\Theta(\boldsymbol{x})_{i j l}}}\\
    \boldsymbol{M}_{i j k}^{\mathrm{adaptive}}= \sum_{k=1}^{3} \lambda_{\mathrm{penalty}}\cdot \boldsymbol{M}_{i j k}^{\mathrm{original}}\\
    \lambda_{\mathrm{penalty}} = \begin{cases} \frac{\sum \boldsymbol{N}^{\mathrm{class}}}{\boldsymbol{N}^{\mathrm{class}}_{ij}};   \boldsymbol{N}^{\mathrm{class}}_{ij} > 0\\ 0;   \boldsymbol{N}^{\mathrm{class}}_{ij}\leq 0\end{cases}   
\end{aligned}
\end{equation}

where, $\Theta(\boldsymbol{x})_{ijk}$ represents the grasp affordance map of the network, $\boldsymbol{\Psi}_{ijk} \in \mathbb{R}^{H \times W \times 3}$ denotes the ground truth maps, $\boldsymbol{M}_{ijk}^{\mathrm{original}}$ represents the mask without adaptive penalty for the three classes, and $\boldsymbol{N}^{\mathrm{class}}$ represents the number of pixels from specific class. To address the label
imbalance problem, we build an adaptive penalty mask that adjust the weights on the pixels of different classes. Experiments confirmed that the
adaptive penalty mask improves the efficiency and stability of the training procedure and imposes more penalties on negative and positive labels. 
During training, an adaptive variable $\lambda_{penalty}$ is modified according to the label-quantity distribution.

The central task of the model-free grasping network is training the network and improving the generalization when the synthetic data differ from reality and uncertainties exist in the sensor data and contact model. To this end, the network is trained with \textbf{sim-to-real transfer}.

During sim-to-real transfer learning, we generated a synthetic grasping dataset with approximately 65k training samples with more than 9.7 million labels from 1,521 different objects. To bridge the gap between the real and synthesized images, we selected the annotated real-world dataset from the Amazon Robotics Challenge 2017, which contains 389 scenes of only 6.2k sparse grasp samples~\cite{zeng2018robotic}. The network is trained on the synthetic grasping dataset and learned on a human-annotated dataset~\cite{zeng2018robotic} with sim-to-real transfer. The network is trained with Adam and stochastic gradient descent optimizers in GeForce GTX 1080Ti. The learning rate and the batch size are 0.04 and 32, respectively. 

\subsection{Grasp Execution System with Grasp Optimizer}
To reduce the complexity of the computations, we discretize the GWS calculation and apply the grasp orientation quantization method in data post-processing. However, these methods increase the uncertainty of grasp orientation. To obtain precise grasp candidates in the presence of sensor noise, we propose a \textbf{grasp optimizer} that uses the force closure metric to improve the orientation accuracies of the final executing grasps (see Appendix~\ref{sec:appendix_grasp_optimizer}).

The final grasp execution system consists of three main parts: the pre-processor, the encoder-decoder network, and the grasp optimizer. The depth image is rotated 16 times and fed into the trained network, which predicts the grasps parallel to the planar table surface and the x-axis of the image coordinate system. Fig.~\ref{fig:pipeline_overview} (c) shows 16 grasp affordance maps predicted by the network. The horizontal grasp-success possibilities of each pixel are indicated in the 16 rotated input images. The grasp optimizer then fine-tunes the grasp orientation in the grasp affordance maps based on the force-closure metric. Finally, the appropriate grasps are filtered through the gripper model.

\section{Experiment}
\label{sec:experiment}
\subsection{Experimental Setup.}
\begin{figure}[htbp]
\centering
    \includegraphics[width=7.5cm]{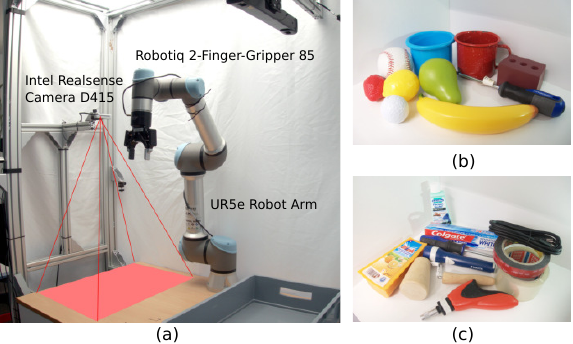}
\caption{(a) Experimental setup: (right) testing set of 20 objects with different geometric features (spheroidal, cuboidal, cuplike, and complicated); (b) 10 known objects in the YCB dataset; (c) 10 unknown household objects.}
\label{fig:experiment_setup_and_test_set}
\end{figure}

Depth images of the scene were observed by an Intel RealSense Depth Camera D415 sensor with a resolution of $640 \times 480$. The physical robotic experiments were executed by a UR5e robot arm mounted with a two-finger gripper Robotiq 85. The camera was placed in an eye-to-hand configuration as shown in Fig.~\ref{fig:experiment_setup_and_test_set}. In the robotic experiments, the arm
was required to lift 10 known objects selected from the YCB dataset~\cite{calli2017yale} and 10 novel household items (see Fig.~\ref{fig:experiment_setup_and_test_set}).

\subsection{Qualitative Study.}
To understand the distribution of the graspable possibilities in our framework, we analyzed the grasp affordance maps with and without our proposed data augmentation. In a grasping-pose analysis, we also investigated the feasibility of the orientation quantization and the corresponding rotation of depth images and label maps. The data augmentation improved the data distribution of network results and reduced the possibility of grasp candidates at the object edges and table surface. Additional details are presented in Appendix~\ref{sec:appendix_qualitative_study}.

\begin{figure}[htbp]
\centering
\includegraphics[width=7.5cm]{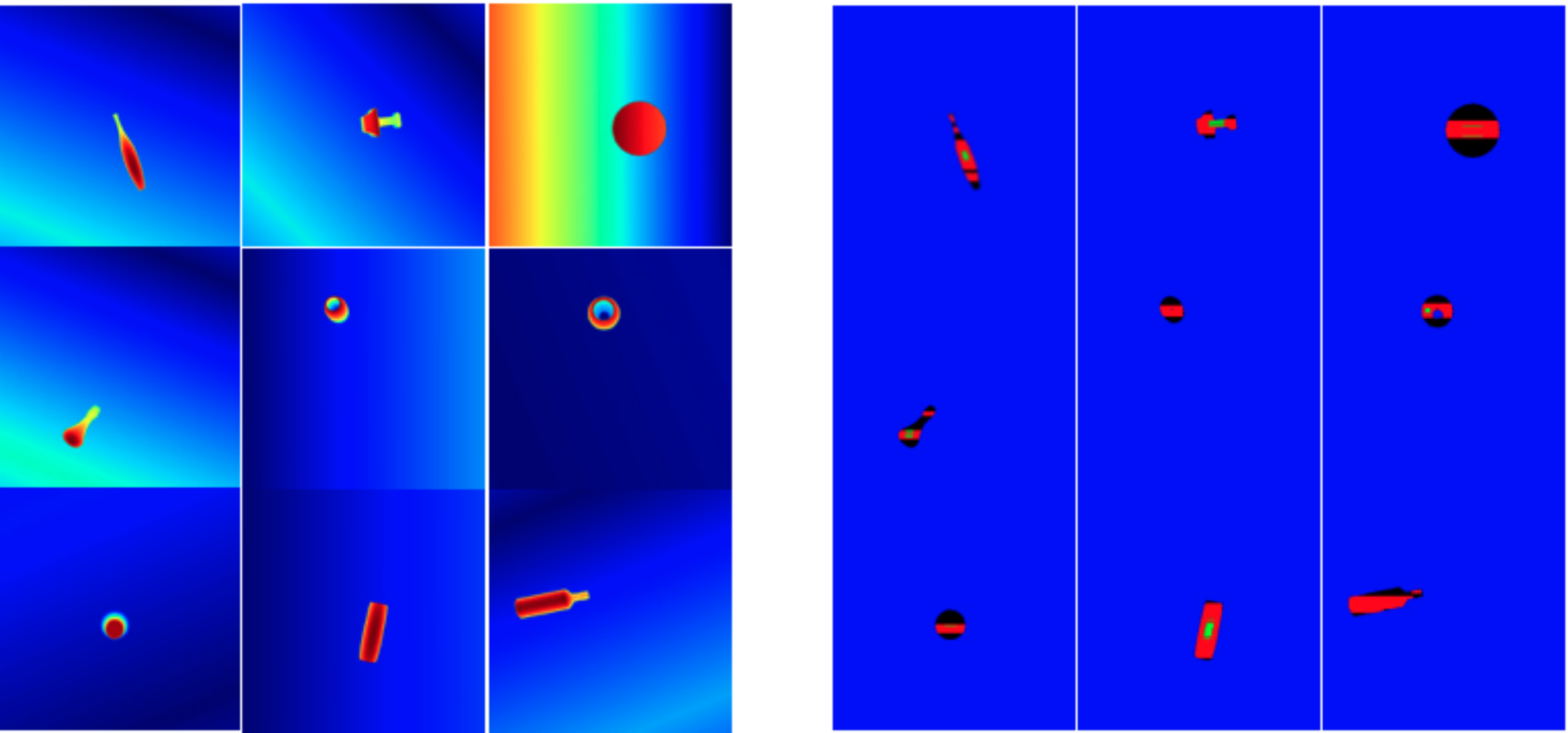}
\caption{Synthetic dataset based on our data augmentation. (left) depth images; (right) corresponding label maps.}
\label{fig:example_synthetic_dataset}
\end{figure}

\subsection{Comparison Experiments.}
After training and evaluating our model-free grasping frameworks, we compared the performances of the frameworks with different structures learned on different training datasets with those of two state-of-the-art methods (Dex-Net 2.0~\cite{mahler2017dex} and MetaGrasp~\cite{2019Metagrasp}). The experimental setup is shown in Fig.~\ref{fig:experiment_setup_and_test_set}. In a trial, two lifts and regrasps are allowed after the initial failure.

\textbf{Evaluation Metric.} As the evaluation metric of grasp success rate, we divided the number of successful grasps by the total number of successful and failed grasps. As the robot sometimes repeats a failure grasp, the grasp was defined as a "failure" when the object grasping failed for the same reason in two lifting operations. Newly failed grasps and the grasping of two objects in one trial were also treated as failures.

\textbf{Synthetic and Real Datasets.} The following diverse robotic grasping datasets were generated for the comparison experiments and sim-to-real transfer learning:
\begin{itemize}
\item Syn-Train: A synthetic training set with over 65k samples, augmented with proposed data augmentation method. Fig.~\ref{fig:example_synthetic_dataset} shows examples of our synthetic dataset.
\item Syn-Real-Train: A training set with synthetic images and 6.2k real-world samples with sparse grasp labels from~\cite{zeng2018robotic}.
\item Real-Train: A training set composed only of real-world samples.
\end{itemize}
% Note 

\begin{table}[htbp]
\centering
\vspace*{-3mm}
	\caption{Outcomes of grasping comparison experiments}
	\label{tab.grasp_comparison}
	\vspace*{-3mm}
\begin{threeparttable}
\begin{tabular}{|c|c|c|c|c|}
\hline
&&
\multicolumn{3}{c|}{ \multirow{1}*{Success Rate (\%)}}\\
\cline{3-5}
 \multirow{-2}{*}{Method} &
 \multirow{-2}{*}{Trials} & YCB-obj & household-obj & multi-obj\\
\hline
DexNet 2.0\tnote{*} & 145 & 88.2 & 83.3 & 80.0 \\
\hline
Metagrasp\tnote{*} & 148 & 67.4 &  \textbf{93.8} & 79.5 \\
\hline
Syn-Train+ &&&&\\
without-GO\tnote{*} & \multirow{-2}*{174} & \multirow{-2}*{80.6} & \multirow{-2}*{75.0} & \multirow{-2}*{60.2} \\
\hline
Syn-Real-Train+&&&&\\
without-GO\tnote{*} & \multirow{-2}*{165} & \multirow{-2}*{85.3} & \multirow{-2}*{75.0} & \multirow{-2}*{62.6} \\
\hline
Syn-Real-Train+&&&&\\
with-GO\tnote{*} & \multirow{-2}*{136} & \multirow{-2}*{ \textbf{90.9}} & \multirow{-2}*{90.9} & \multirow{-2}*{ \textbf{85.7}} \\
\hline
\end{tabular}
 \begin{tablenotes}
        \footnotesize
        \item[*] Success rates are based upon testing using a UR5e physical robot. The 10 known objects are selected from the YCB dataset and the novel testing set consists of 10 household objects.  %
      \end{tablenotes}
    \end{threeparttable}
    \vspace*{-3mm}
\end{table}

\textbf{Model-free Grasping Frameworks.}
Besides the three datasets described above, we generated the following framework structures for the comparison experiments (see Table~\ref{tab.grasp_comparison}).
\begin{itemize}
\item Networked-without-GO: A networked model-free framework that directly executes grasping following the grasp affordance map without the grasp optimizer.
\item Networked-with-GO: A networked model-free grasping framework that improves the orientation accuracy of the grasp by applying the final grasp optimizer to the grasp affordance map. The optimizer considers the force-closure metric and implements the gripper model.
\end{itemize}

\textbf{Experimental scenarios.} The following scenarios were constructed for the comparison experiments:

\begin{itemize}
\item Scenarios with individual objects: The grasped object is placed on the table in a random position.
\item Cluttered scenarios with multiple objects: Randomly selected objects are placed on the table. 
\item Task-oriented scenarios with multiple data cables: Various data cables are placed on the table mimicking the model-free grasping conditions of industrial applications.
\end{itemize}

\section{Results}\label{sec:results}
\subsection{Real-world Grasping of Known and Novel Objects}
The statistical results of the comparison experiments are provided in Table~\ref{tab.grasp_comparison}. Evidently, the highest-performing pipeline was the third version of our method (Syn-Real-Train+Networked-with-GO), indicating that sim-to-real transfer learning with dense grasp labels and the grasp optimizer improved the performance of grasping unknown objects. A network trained on our synthetic dataset with dense grasp labels also achieved good performance on real-world data. In addition, the grasp optimizer reduced the grasp-quantization errors in the grasp orientation and improved the capability of grasping objects with complex structures. The domain randomization method guaranteed the performance of grasping of objects with different sizes.

The Metagrasp pipeline achieved a high success rate when grasping single unknown objects with different shapes of normal size, but its performance degraded
when grasping small objects such as golf balls and strawberries, possibly
because Metagrasp data are generated without domain randomization. In addition, examining the affordance map of Metagrasp, it appears that the best grasp position located on the table surface owing to sparsity of the grasp pixel labels.

DexNet 2.0 has limited ability for grasping small objects despite the large-scale dataset generated for DexNet 2.0. Dex-Net 2.0 also demonstrated limited performance for objects with multiple local structures, perhaps because the method is constrained to local feature-based grasp representation for real-world applications.

\subsection{Task-oriented Grasping by Physical Robot}
Robotic grasping of complex objects is recognized as a difficult task in research and industry. In this experiment, we tested our framework in a scenario of multiple data cables with complicated structures. In 13 trials, the success rate of clean grasping of data cables was 92.31\%. Judging from this result, our robotic grasping framework is a robust and effective solution for unknown object grasping in complicated scenarios (see Appendix~\ref{task-oriented real-world robotic grasping}). 
\section{Conclusion AND Discussion}
\label{sec:conculsion}
We proposed a novel data augmentation method that generates datasets for robotic grasping. The method resolves the problem of sparse grasping pixel labels in current synthetic and human-annotated datasets. Specifically, the labels are evaluated with the force-closure and grasp measure metrics, which improve the reliability of the grasp dataset. Using our data generation methods with domain randomization, we generated a large-scale grasping dataset with over 65k synthetic samples. We also presented an adaptive loss function for robust training on label-imbalanced datasets and trained an end-to-end grasping affordance network with sim-to-real transfer. Finally, we combined the affordance map generated by our grasping framework with a grasp optimizer that reduces the orientation error caused by discretization and quantization of the grasp orientation. In physical experiments, a robot programmed with our approach successfully grasped known or unknown individual objects of different shapes. The success rate was 90.91\%. In multi-object scenarios, the success rate dropped but remained high at 85.71\%. Our pipeline was also evaluated in a challenging task-oriented robotic grasping environment, in which several complicated data cables were placed on the table. Under our framework, the robot grasped the data cables with a success rate of 92.31\%.

In both quantitative studies and real-world robotic experiments, our framework outperformed two previously published methods~\cite{mahler2017dex,2019Metagrasp}. In particular, it achieved the highest performance in known and unknown object robotic grasping in complex scenarios. It also performed well in a challenging and specific industrial activity.

In future work, we will investigate a dexterous manipulation framework with humanoid robots (five-finger hands and robot arms) combining current data generation methods with cross-modal fusion of different modalities.

%%%%%%%%%%%%%%%%%%%%%%%%%%%%%%%%%%%%%%%%
\section*{ACKNOWLEDGMENT}
This research has received funding from the German Research Foundation (DFG) and the National Science Foundation of China (NSFC) in project Crossmodal Learning, DFG TRR-169/NSFC 61621136008, partially supported by European projects H2020 STEP2DYNA (691154) and ULTRACEPT (778602). We thank Max-Heinrich Laves and Prof. Dr. Tobias Ortmaier for their valuable feedback during the discussion of the experimental protocol.
%%%%%%%%%%%%%%%%%%%%%%%%%%%%%%%%%%%%%%%%%%%%%%%%%%%%%%%%%%%%%%%%%%%%%%%%%%%%%%%%

\bibliographystyle{IEEEtran}
{\scriptsize
\vspace{0.01 cm}
\bibliography{main}
} 
%%%%%%%%%%%%%%%%%%%%%%%%%%%%%%%%%%%%%%%%%%%%%%%%%%%%%%%%%%%%%%%%%%%%%%%%%%%%%%%%
\section*{APPENDIX}
\subsection{Qualitative Study}\label{sec:appendix_qualitative_study}
\subsubsection{Statistical Horizontal Grasping Analysis}\label{sec:appendix_statistical_analysis}
\begin{figure}[H]
\centering
    \includegraphics[width=6.0cm]{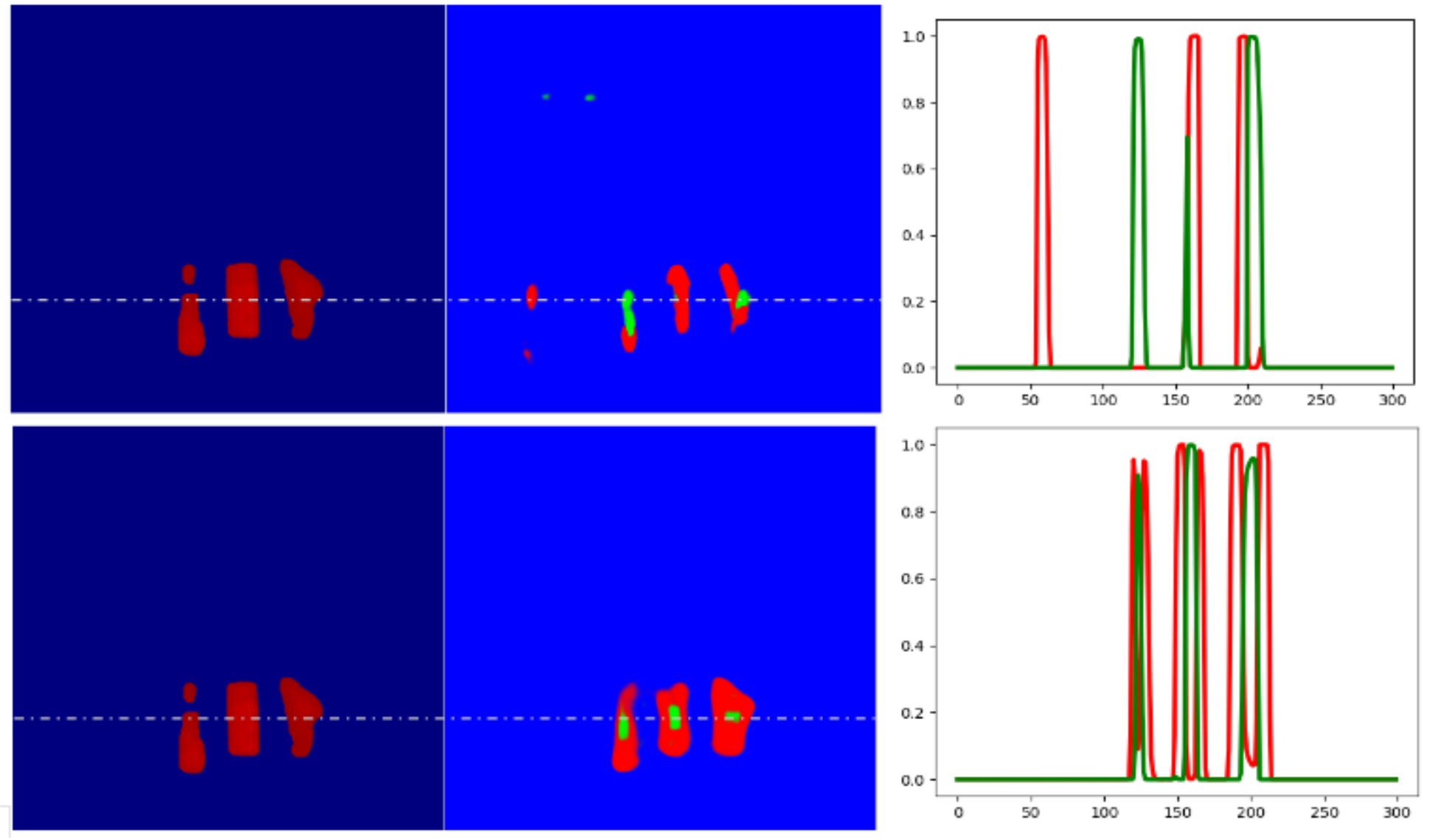}
\centering
\caption{Statistical horizontal grasping analysis without and with data augmentation.  (left) Depth images, (middle) horizontal grasp affordance maps, (right) grasp probability distributions at the marked positions (red: negative grasp possibility, green: positive grasp possibility). In the top and bottom rows, training was performed on Real-Train dataset without data augmentation and Syn-Train dataset with data augmentation. }
\label{fig:inference_map}
\end{figure}

Fig. \ref{fig:inference_map} depicts the inference inputs and affordance maps for horizontal grasping trained on the Syn-Train dataset with proposed data augmentation, the Real-Train dataset without data augmentation, and the Syn-Real-Train dataset. The red and green regions in the affordance maps represent the negative and positive grasp central positions, respectively, and the blue regions are the background. The right panels of Fig.~\ref{fig:inference_map} present the grasping probability statistics at the marked positions in the affordance maps trained on Syn-Train dataset and Syn-Real-Train dataset. The red and green curves are the probability distributions of the negative and positive grasps, respectively.

Observing the result in Fig.~\ref{fig:inference_map}, our augmentation improved the quality of the affordance map and the grasp probability distribution. The network trained on the non-augmented dataset obtained a poor inference result in the region of unlabeled grasp classes. The final executed grasp may locate on the object edge, which is a negative grasping action in real robot system; accordingly, the uncertainty of the trained CNN increased without the data augmentation. Our data augmentation method reduced the uncertainty of the trained network and the statistical method accurately and rapidly increased the number of labels.

From the affordance map and horizontal-grasp probability distributions in Fig.~\ref{fig:inference_map}, it appears that the most probable grasps were
horizontally oriented grasps centralized at the the mid-point
of each object. These results prove that the trained network successfully detected horizontal force-closure grasps.
\subsubsection{Grasping Pose Analysis}
\begin{figure}[H]
\centering
\includegraphics[width=8cm]{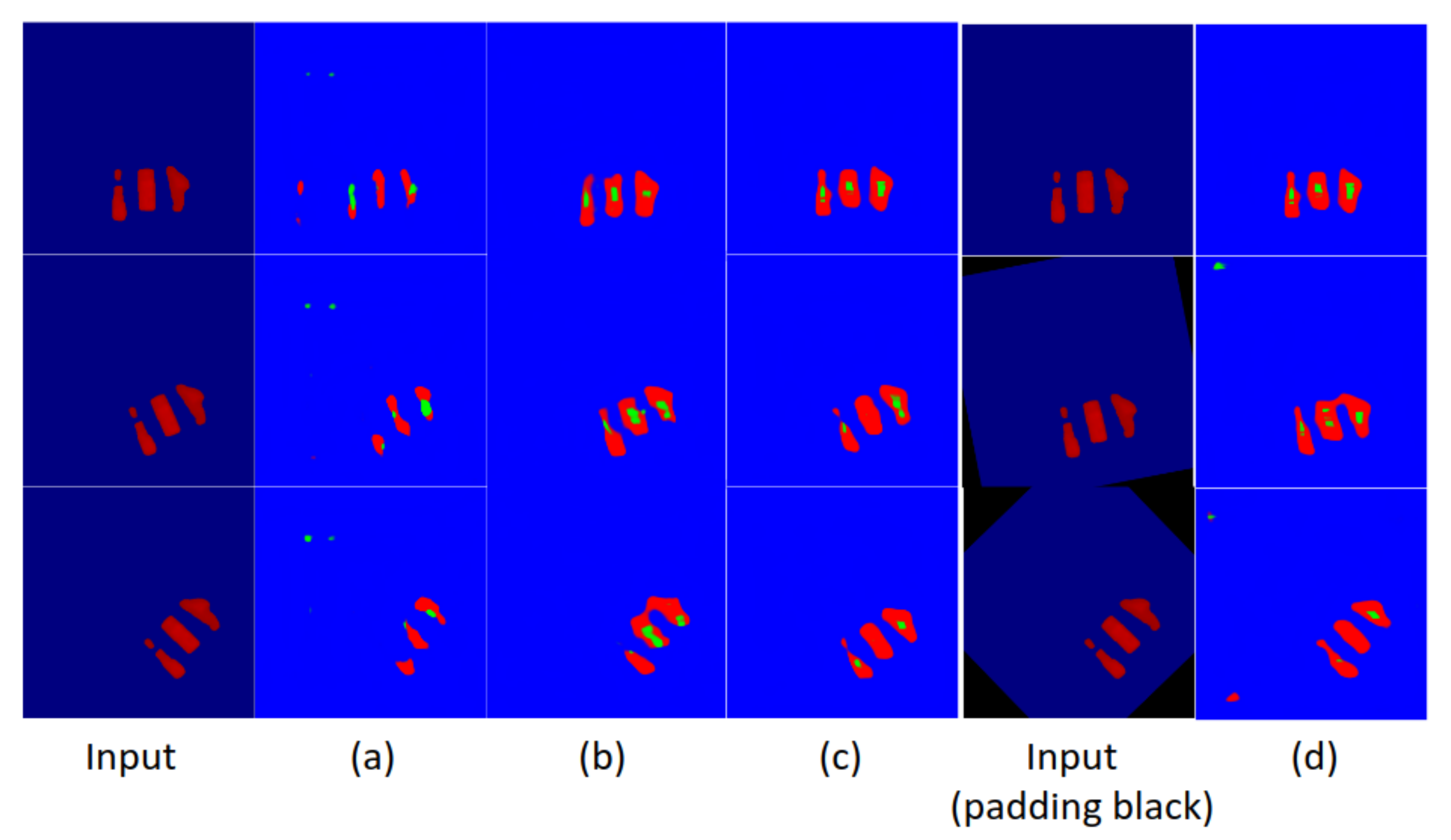}
\centering
\caption{Grasping pose analysis. The rotated encoded input image was fed into a trained network and the affordance maps were estimated. Redand green pixels are the central points of positive and negative grasps, respectively, and blue
pixels are background. Shown are the results of (a) the network trained on a dataset without proposed data augmentation, (b) the network trained on the Syn-Train dataset, (c) the network trained pon the Syn-Real-Train dataset with sim-to-real-transfer. Panels (c) and (d) compare the results of different padding colors during the rotation operations. (c) and (d) use the same training set. (a)~(c): padding with adaptive color, (d): black padding.}
\label{fig:inference_map_comparison_train_set_various_pad}
\end{figure}

Fig. \ref{fig:inference_map_comparison_train_set_various_pad} compares the grasp affordance maps generated by networks trained on different training sets and with various padding colors during the rotation operations before the input enters the network. The network trained on the dataset without proposed augmentation misclassified some pixels in the table and positive grasps were predicted
at some edge pixels. This type of uncertainty significantly degrades the performance of real-world robotic grasping. For this reason, the non-augmented synthetic dataset was not applied in the robotic experiments. Panels (b) and (c) of Fig. \ref{fig:inference_map_comparison_train_set_various_pad} respectively show the inference results of the networks trained on synthetic data alone (Syn-Train dataset) and on the Syn-Real-Train dataset, which contains synthetic data and a small around of real data. Our Syn-Train dataset with over 65k of synthetic data achieved good results in scenes with multiple objects. Meanwhile, the Syn-Real-Train dataset improved the grasp orientation results during rotations of the input image. We also tested an adaptive depth-based padding method for the rotation operation and investigated the results of input images with adaptively colored padding and black padding (as used in \cite{2019Metagrasp} and \cite{zeng2018robotic}). When the black padding was used (Fig. \ref{fig:inference_map_comparison_train_set_various_pad} (d)), a few pixels on the table were erroneously labeled in green or red. Such errors will reduce the stability and ability of the entire framework.
\subsection{Grasp Optimizer with Analytical Methods}
\label{sec:appendix_grasp_optimizer}
The grasp optimizer in our robotic grasping framework selects the grasps with higher possibility as candidates of the final grasp. The prediction values of the positive and negative grasp classes are calculated from the grasp affordance maps.
\begin{figure}[ht]
\centering
    \includegraphics[width=8cm]{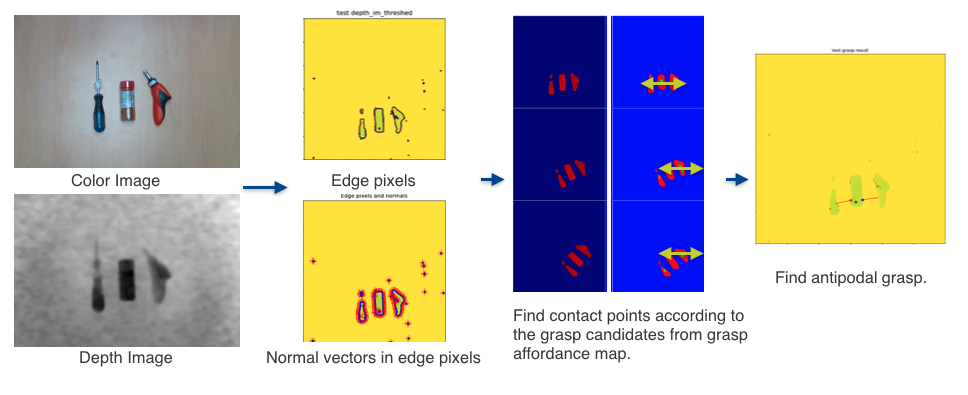}
\centering
\caption{Pipeline of the grasp optimizer for finding the optimal grasp candidate.}
\label{fig:grasp_optimization_pipline}
\end{figure}
Fig. \ref{fig:grasp_optimization_pipline} shows the whole pipeline of the grasp optimizer. First, the edge pixels are detected as gradients in the depth image. Next, the contact points of each candidate are estimated based on the grasp central point and grasp direction planned in the grasp affordance map. After obtaining two contact points, a grasp width filter is applied. A grasp candidate is considered as a valid candidate if the distance between the contact points is smaller than the maximal gripper width. Finally, the valid candidate is adjusted depending on the antipodal or force-closure grasp metric. 
%The 
To improve the success rate of grasping and to optimize the grasp direction, the normal direction of the contact points is set in the direction of the final executed grasp (see Figure \ref{fig:grasp_optimization_pipline}). The grasp optimizer increases the robustness of the robotic grasping framework and reduces the number of failures caused by external factors, such as collisions during grasping.
%Th
\subsection{Task-oriented Real-world Robotic Grasping}
\label{task-oriented real-world robotic grasping}
\begin{figure}[H]
\centering
\includegraphics[width=9cm]{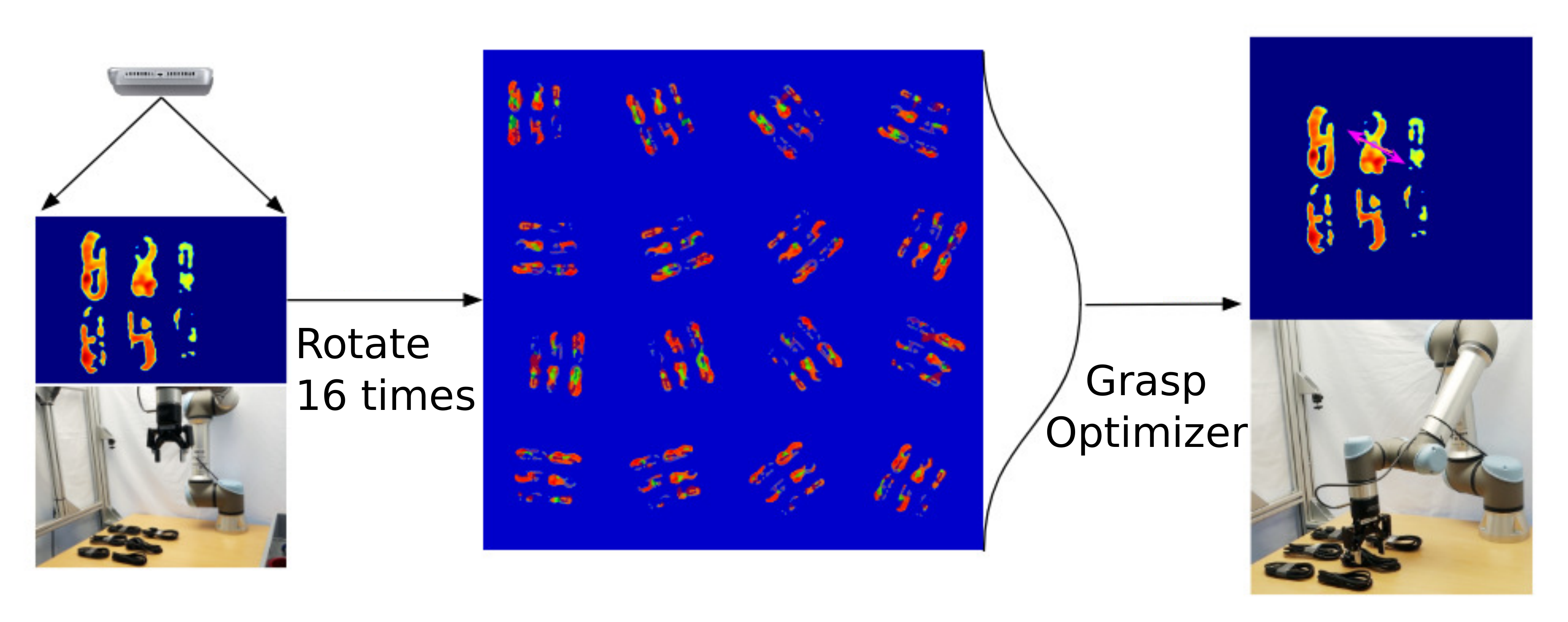}
\centering
\caption={Pipeline of the task-oriented robotic grasping experiment using our grasping framework. From left to right: Depth image, affordance map, final grasp. The right panel also shows the finally executed grasp candidate.}
\label{fig:task_oriented_experiment_pipline}
\end{figure}

% %%%%%%%%%%%%%%%%%%%%%%%%%%%%%%%%%%%%%%%%%%%%%%%%%%%%%%%%%%%%%%%%%%%%%%%%%%%%%%%%

\end{document}